\documentclass[11pt]{article}

\usepackage[preprint]{acl}
\usepackage{amsmath} 
\usepackage{times}
\usepackage{latexsym}
\usepackage{amsthm}
\newtheorem{definition}{Definition}[section]

\usepackage[T1]{fontenc}
\usepackage{tikz}
\usetikzlibrary{shadows.blur}
\usepackage{forest}
\usetikzlibrary{arrows.meta,positioning,shapes.misc}
\usetikzlibrary{arrows.meta} 
\usepackage{tikz-qtree}
\usetikzlibrary{trees} 

\usepackage[utf8]{inputenc}

\usepackage{microtype}

\usepackage{inconsolata}
\usepackage{amsmath,amssymb}
\usepackage{graphicx}
\usepackage{booktabs}
\usepackage{amsmath,amsfonts,bm}

\def\eqref#1{equation~\ref{#1}}

\def\1{\bm{1}}

\DeclareMathAlphabet{\mathsfit}{\encodingdefault}{\sfdefault}{m}{sl}
\SetMathAlphabet{\mathsfit}{bold}{\encodingdefault}{\sfdefault}{bx}{n}

\title{From Efficiency to Adaptivity: A Deeper Look at Adaptive Reasoning in Large Language Models}

\author{
  Chao Wu \\
  University at Buffalo\\
  \texttt{cwu64@buffalo.edu} \\
  \And
  Baoheng Li \\
  University at Buffalo\\
  \texttt{baohengl@buffalo.edu} 
  \AND
  Mingchen Gao \\
  University at Buffalo\\
  \texttt{mgao8@buffalo.edu}
  \And
  Yu Tian \\
  University of Central Florida \\
  \texttt{yu.tian2@ucf.edu}
  \And
  Zhenyi Wang \\
  University of Central Florida \\
  \texttt{zhenyi.wang@ucf.efu} 
}

\begin{document}
\maketitle
\begin{abstract}
Recent advances in large language models (LLMs) have made reasoning a central benchmark for evaluating intelligence. While prior surveys focus on efficiency by examining how to shorten reasoning chains or reduce computation, this view overlooks a fundamental challenge: current LLMs apply uniform reasoning strategies regardless of task complexity, generating long traces for trivial problems while failing to extend reasoning for difficult tasks. This survey reframes reasoning through the lens of {adaptivity}: the capability to allocate reasoning effort based on input characteristics such as difficulty and uncertainty. We make three contributions. First, we formalize deductive, inductive, and abductive reasoning within the LLM context, connecting these classical cognitive paradigms with their algorithmic realizations. Second, we formalize adaptive reasoning as a control-augmented policy optimization problem balancing task performance with computational cost, distinguishing learned policies from inference-time control mechanisms. Third, we propose a systematic taxonomy organizing existing methods into training-based approaches that internalize adaptivity through reinforcement learning, supervised fine-tuning, and learned controllers, and training-free approaches that achieve adaptivity through prompt conditioning, feedback-driven halting, and modular composition. This framework clarifies how different mechanisms realize adaptive reasoning in practice and enables systematic comparison across diverse strategies. We conclude by identifying open challenges in self-evaluation, meta-reasoning, and human-aligned reasoning control.
\end{abstract}

\section{Introduction}
Reasoning is a core aspect of intelligence, allowing humans to draw inferences from evidence, rules, and prior knowledge to make systematic decisions. In the context of large language models (LLMs), reasoning has become a central benchmark of progress, not only for solving complex tasks such as mathematics, programming, and scientific discovery, but also for demonstrating abilities beyond pattern recognition. As LLMs scale in capability, understanding how they reason and how to control their reasoning processes has emerged as a fundamental research question with both theoretical and practical implications.

Existing surveys on LLM reasoning have largely focused on efficiency: how to shorten reasoning chains, avoid overthinking, and reduce computational cost. While efficiency is an important practical concern, it does not fully capture the broader challenge. Current LLMs often adopt a one-size-fits-all reasoning strategy, generating long reasoning traces even for trivial problems while failing to extend reasoning when faced with difficult tasks. What is missing is not merely efficiency, but adaptivity, which is the ability to flexibly adjust reasoning strategies based on context, difficulty, and uncertainty. We define adaptive reasoning in LLMs as the capability to allocate reasoning effort based on input characteristics such as difficulty and uncertainty. This adaptivity can be realized through learned policies that internalize allocation strategies during training, or through dynamic control mechanisms applied at inference time. Unlike efficiency-centric approaches that uniformly constrain reasoning length, adaptive reasoning emphasizes input-dependent resource allocation: answering easy questions quickly while allocating deliberate reasoning effort to harder or unfamiliar problems, thereby mirroring the natural scaling of human cognitive effort with task demands.

To realize adaptive reasoning, LLMs must flexibly employ different reasoning strategies depending on task demands. Reasoning in LLMs encompasses diverse cognitive patterns, among which three classical paradigms—inductive, deductive, and abductive reasoning—provide a useful conceptual lens. Inductive reasoning generalizes from observed examples, as seen in in-context learning \cite{brown2020language}; deductive reasoning applies explicit logical rules to derive valid conclusions, underlying structured chain-of-thought \cite{wei2022chain} and program-of-thought \cite{chen2022program} frameworks; abductive reasoning seeks plausible explanations for observed outcomes. However, rather than organizing methods by reasoning type, this survey focuses on how adaptivity is implemented in practice. We propose a systematic taxonomy distinguishing training-based approaches, which internalize adaptive allocation strategies through reinforcement learning, supervised fine-tuning, and learned controllers, from training-free approaches, which achieve adaptivity through prompt conditioning, feedback-driven halting, and modular composition at inference time.
Our contributions are threefold:
\begin{itemize}
\item We reframe LLM reasoning research through the lens of {adaptive reasoning}, moving beyond efficiency-centric approaches to emphasize input-dependent resource allocation. We define adaptivity as the capability to allocate reasoning effort based on task difficulty and uncertainty, realized through either learned policies or dynamic inference control.
\item We formalize deductive, inductive, and abductive reasoning within the LLM context, providing clear definitions that connect these classical cognitive paradigms with their algorithmic realizations in contemporary LLM systems.
\item We propose a systematic taxonomy organizing existing methods by implementation mechanism, distinguishing training-based approaches that learn adaptive policies from training-free approaches that apply dynamic control at inference and enabling systematic comparison across diverse adaptive reasoning strategies.
\end{itemize}
\section{Background and Foundations}
\subsection{What is reasoning in LLMs?}

Reasoning is a cornerstone of human intelligence, enabling inference and decision-making from evidence and rules. Following \citet{huang-chang-2023-towards} and \citet{10.1145/3729218}, reasoning can be viewed as a \textit{logical and systematic process} that uses prior knowledge and contextual evidence to arrive at valid conclusions or hypotheses. Within the context of LLMs, reasoning refers to the model's capacity to generate and verify intermediate representations that connect an input prompt $\mathbf{x}$ to an output $y$ through a structured inference procedure.

We formalize reasoning from two complementary perspectives: (1) as a latent-variable conditional generation process (\S\ref{conditional gen}), and (2) as a resource-bounded optimization problem (\S\ref{optimize}). These formalizations provide the mathematical foundation for understanding adaptive reasoning mechanisms.

\subsubsection{Reasoning as Conditional Generation}
\label{conditional gen}
Following \citet{bandyopadhyay2025thinkingmachinessurveyllm}, A language model defines a conditional probability distribution over output sequences:
\begin{equation}
p_{\boldsymbol{\theta}}(\mathbf{y}\mid \mathbf{x})
= \prod_{n=1}^{N} p_{\boldsymbol{\theta}}(y_n \mid \mathbf{x}, y_{<n}),
\end{equation}
where $\boldsymbol{\theta}$ denotes model parameters.

Extending this formulation, \citet{phan2023training} formalize reasoning 
as a latent-variable conditional generation process. 
Specifically, the model can be viewed as first generating a latent reasoning 
trace $\mathbf{r}$, which is an unobserved sequence of intermediate thought step, and 
then producing the final output $\mathbf{y}$ conditioned on both $\mathbf{x}$ 
and $\mathbf{r}$. This leads to the marginal likelihood:
\begin{equation}
\log p_{\boldsymbol{\theta}}(\mathbf{y}\mid \mathbf{x})
= \log \sum_{\mathbf{r}} 
p_{\boldsymbol{\theta}}(\mathbf{r}\mid \mathbf{x})\,
p_{\boldsymbol{\theta}}(\mathbf{y}\mid \mathbf{r},\mathbf{x}).
\label{conditional equation}
\end{equation}
Under this latent-variable perspective, reasoning is formalized as 
marginalizing over possible reasoning trajectories that mediate between 
the input $\mathbf{x}$ and the final output $\mathbf{y}$. 
This view conceptually unifies reasoning and conditional text generation,
suggesting that the ``thought'' process of an LLM can be understood as sampling 
from an implicit distribution over intermediate reasoning states.

\subsubsection{Reasoning as Resource-Bounded Optimization}
\label{optimize}
Following \citet{alomrani2025reasoningbudgetsurveyadaptive}, 
test-time reasoning can be viewed as a trade-off between computation and performance. 
Two equivalent optimization views are commonly considered: 
a constrained form ($L_1$) that fixes the compute budget, 
and a penalized form ($L_2$) that adaptively balances accuracy and efficiency. 
We adopt the $L_2$ formulation, which naturally aligns with the idea of adaptive reasoning:
\begin{equation}
\max_{\mathbf{r}\in\mathcal{R}}\;
\mathcal{P}(\mathbf{r}, \mathbf{x})
+ \alpha\,\mathcal{E}(\mathbf{r}, \mathbf{x}),
\end{equation}
where $\mathcal{P}$ measures task performance, 
$\mathcal{E}$ quantifies computational efficiency 
(e.g., the inverse of compute cost), 
and $\alpha$ controls their trade-off. 
Although $L_2$ removes the explicit resource constraint in $L_1$, 
it serves as its \emph{Lagrangian relaxation}, 
implicitly preserving the same budgeted principle through the weighted efficiency term. 
This formulation highlights reasoning as an adaptive optimization process 
under limited computational resources.

\subsubsection{Limitations of Efficient Reasoning}

\label{sec:limitations-efficiency}
LLMs often exhibit an \emph{overthinking} phenomenon, producing unnecessarily long chains of thought (CoT) for simple problems, thereby increasing compute without improving accuracy
\citep{Sui2025StopOverthinking,alomrani2025reasoningbudgetsurveyadaptive}.
Efficiency-oriented methods attempt to limit reasoning tokens, steps, or decoding paths to reduce test-time compute. 
However, such \textit{static control} allocates identical budgets to all inputs, regardless of difficulty or confidence, 
leading to redundant reasoning on easy cases and insufficient reasoning on complex ones.

Empirical studies confirm that fixed-length CoT models often waste computation on easy inputs while failing on hard ones \citep{Sui2025StopOverthinking}. 
The key limitation is that computation is treated as an external constraint rather than a self-regulated process: 
intelligent reasoning should dynamically decide how much to think based on uncertainty and task complexity.

To address these limitations, this survey proposes adaptive reasoning as a unified framework that (1) characterizes reasoning through broader cognitive paradigms (§\ref{sec:formal-reasoning}), and (2) formalizes adaptivity as a policy optimization problem (§\ref{sec:transfer-to-adaptive}), distinguishing methods that internalize adaptivity through training from those that apply external control at inference.

\subsection{Forms of Reasoning}
\label{sec:formal-reasoning}
We formalize three classical reasoning paradigms that LLMs may employ. Let $\mathbf{x}$ denote the input, $\mathbf{y}$ the output, and $\mathbf{r} = (r_1, \ldots, r_K)$ the intermediate reasoning steps. Each paradigm defines how the model constructs and evaluates reasoning trajectories.

\paragraph{Setup.}
A reasoning process generates an intermediate trajectory 
$\mathbf{r} = (r_1, \ldots, r_K)$ that mediates between input $\mathbf{x}$ 
and output $\mathbf{y}$, where each step $r_k$ is autoregressively conditioned on $(\mathbf{x}, r_{<k})$.
We call a reasoning episode $R_{\boldsymbol{\theta}}(\mathbf{x})=(\mathbf{r},\mathbf{y})$ \emph{valid}
if it satisfies (1) \textbf{Causal coherence} — each step $r_k$ depends only on $\mathbf{x}$ and previous steps, ensuring a forward information flow;
(2) \textbf{Goal consistency} — the reasoning trajectory $\mathbf{r}$ produces an answer $\mathbf{y}$ that faithfully and correctly fulfills the task objective, following the principles of faithful and step-wise reasoning \citep{lee2025evaluating,lyu2023faithful}

\begin{definition}[\textbf{Deductive Reasoning}]
A reasoning process is \emph{deductive} if each intermediate step $r_k$ 
logically follows from the input $\mathbf{x}$ and prior steps, and the 
final answer $y$ is necessarily entailed:

\begin{equation}
\begin{aligned}
&\mathbf{x} \cup \{r_1,\dots,r_{k-1}\} \models r_k, 
\quad &&\forall k \in \{1,\dots,K\},\\[3pt]
&\mathbf{x} \cup \mathbf{r} \models y. &&
\end{aligned}
\label{eq:deductive}
\end{equation}
\end{definition}
Here, the symbol ``$\models$'' denotes logical entailment: 
a set of premises $A$ entails a conclusion $B$ 
($A \models B$) if every model that satisfies all formulas in $A$ 
also satisfies $B$. 

This ensures that the reasoning trajectory preserves logical validity 
at every step~\citep{ling2023deductive}. In LLMs, deductive reasoning 
is approximated through verifiable chain-of-thought generation, where 
each step can be checked for logical entailment from previous 
statements~\citep{ling2023deductive,SealsShalin2024Deductive,Xia2025FormalLogic}.

\begin{definition}[\textbf{Inductive Reasoning}]
\label{def:inductive}
Let $E = \{(\mathbf{x}_i, y_i)\}_{i=1}^K$ be observed examples, 
$\mathcal{H}$ a hypothesis space $h: \mathcal{X} \to \mathcal{Y}$,
and $C: \mathcal{H} \to \mathbb{R}_{\ge0}$ a complexity measure. 
An inductive reasoning process infers:
\begin{equation}
\label{eq:inductive}
h^* = \arg\min_{h \in \mathcal{H} : h(\mathbf{x}_i)=y_i \,\forall i} C(h),
\end{equation}
the simplest hypothesis consistent with all observations.
\end{definition}

This formalizes \emph{Occam's Razor}~\citep{Baker2022-BAKSSE-2} that preferring 
hypotheses that explain data with minimal complexity. 
When $C(h) = -\log p(h)$, this reduces to Maximum A Posteriori (MAP) 
estimation under Bayesian induction.
In LLMs, inductive reasoning manifests primarily in \emph{in-context 
learning} (ICL)~\citep{brown2020language}, where models extrapolate implicit 
rules from few-shot demonstrations to unseen cases~\citep{Ye2025ICLAttribution}.
However, empirical studies show that LLMs do not always follow simplicity 
principles in practice~\citep{SunSaparov2025Occam}.

\begin{definition}[\textbf{Abductive Reasoning}]
\label{def:abductive}
Given observations $O$ and background knowledge $\mathcal{B}$,
abductive reasoning generates the most plausible explanatory 
hypothesis $H^*$ from a candidate space $\mathcal{H}$:
\begin{equation}
\label{eq:abductive}
H^{*} = \arg\max_{H \in \mathcal{H}} p_\theta(H \mid O, \mathcal{B})
\quad\text{s.t.}\quad \mathcal{B} \cup H \vdash O,
\end{equation}

where $p_\theta(H \mid O, \mathcal{B})$ quantifies the model's belief in the plausibility 
of each hypothesis given the observation and prior knowledge. \end{definition}
Here, the symbol ``$\vdash$'' denotes \emph{syntactic entailment}---that is, 
the observation $O$ can be formally derived from the combined premises 
$\mathcal{B}$ and $H$ according to a logical inference system.

This instantiates \emph{inference to the best explanation}~\citep{Baker2022-BAKSSE-2}:
among hypotheses that entail observation $O$ when combined with background knowledge 
$\mathcal{B}$, abduction selects the most probable one.
In LLMs, abductive reasoning is realized through conditional generation where the 
model samples from $p_\theta(H \mid O)$ to produce plausible explanations.
Benchmarks such as Abductive~\citep{bhagavatula2020abductivecommonsensereasoning} 
and INABHYD~\citep{SunSaparov2025Occam} evaluate this ability by measuring 
explanation coherence and plausibility.

\subsection{Adaptive Reasoning}
\label{sec:transfer-to-adaptive}

Formally, we characterize adaptive reasoning through a {control-augmented policy} 
$\pi_{\boldsymbol{\theta}}(\cdot|\mathbf{x};\boldsymbol{\phi}(\mathbf{x}))$ that maps inputs $\mathbf{x}$ 
to distributions over reasoning trajectories $\boldsymbol{r}$, 
where $\boldsymbol{\theta}$ denotes the base model parameters (e.g., LLM weights) 
and $\boldsymbol{\phi}$ the adaptive control function regulating reasoning behavior.

Adaptive reasoning optimizes the control policy to balance task performance $\mathcal{P}(\boldsymbol{r}, \mathbf{x})$ and computational cost $\mathcal{C}(\boldsymbol{r}, \mathbf{x})$:
\begin{equation}
\label{eq:new adapt}
\max_{\ \boldsymbol{\phi}\in\Phi}\;
\mathbb{E}_{\mathbf{x}\sim\mathcal{D},\, \boldsymbol{r}\sim \pi_{\boldsymbol{\theta}}(\cdot\,|\,\mathbf{x};\,\boldsymbol{\phi}(\mathbf{x}))}
\big[\,\mathcal{P}(\boldsymbol{r},\mathbf{x}) \;-\; \lambda\,\mathcal{C}(\boldsymbol{r},\mathbf{x})\,\big].
\end{equation}
where $\lambda$ controls the cost penalty and $\mathcal{D}$ is the data distribution.
The optimization over $\boldsymbol{\phi} \in \Phi$ captures two complementary approaches to adaptive reasoning. In \textbf{training-based methods} (\S\ref{sec:training-based}), $\boldsymbol{\phi}$ is learned during training, for instance, IBPO \citep{Yu2025IBPO} trains a neural network to predict reasoning budgets, while C3oT \citep{kang2025c3ot} learns to generate short or long reasoning traces conditionally. In \textbf{training-free methods} (\S\ref{sec:training-free}), $\boldsymbol{\phi}$ represents fixed inference-time control such as entropy-based halting \citep{laaouach2025halt} or adaptive sampling thresholds, applied without parameter updates. This unified formulation clarifies how adaptivity arises from either learned policies or inference-time heuristics, both of which modulate $\pi_{\boldsymbol{\theta}}$ to allocate reasoning effort based on input characteristics.

Unlike static efficiency methods that apply uniform compute budgets to all inputs, this framework enables {input-dependent resource allocation} by conditioning reasoning on $\boldsymbol{\phi}(\mathbf{x})$, which adapts to task difficulty and uncertainty. Recent work further shows that such adaptivity can emerge even without reinforcement learning: 
\citet{karan2025reasoning} demonstrates that inference-time adaptive sampling alone 
can elicit strong reasoning behaviors from base LLMs, 
highlighting that reasoning adaptivity may arise naturally from sampling dynamics 
rather than explicit policy optimization.

\paragraph{Adaptive mechanisms in practice.}

Adaptive mechanisms in practice. Existing work realizes adaptivity in two ways, corresponding to whether the policy $\pi_{\boldsymbol{\theta}}$ is optimized during 
training or fixed at inference, shown in our taxonomy(§\ref{sec:taxonomy}).

\textbf{(1) Training-free adaptivity.}
Adaptivity can emerge without retraining the model, by dynamically
regulating reasoning length or sampling during inference.
\citet{laaouach2025halt} halt generation when the stepwise entropy
\(H_i=-\!\sum_a p_i(a)\log p_i(a)\) drops below a threshold~$\theta$.
\citet{aggarwal2023letssamplestepstep} use a Dirichlet prior over
samples to detect when majority confidence exceeds \(C_{\text{thresh}}\),
stopping further decoding.
\citet{Han2025TALE} estimates minimal token budgets per instance and
predicts per-sample cost as
\(\text{Cost}=\frac{1}{N}\sum_{i=1}^N T(M(x_i))\).
These methods achieve instance-level compute control through
entropy, confidence, or budget signals.

\textbf{(2) Training-based adaptivity.}
Other approaches learn a reasoning policy that internalizes computation
allocation during optimization.
\citet{Yu2025IBPO} formulate inference-aware optimization
with explicit budget constraints, where the policy is trained 
to maximize reward while satisfying correctness thresholds 
that implicitly controls compute allocation.
Such training-based strategies generalize TTC to the learning phase,
yielding policies that adapt depth and effort before deployment.

\textbf{Key properties.} ~\eqref{eq:new adapt} 
directly encodes three key properties of adaptive reasoning: 

\textbf{(1)~Input-dependent allocation}—$\pi_\theta(\cdot|\mathbf{x})$ 
conditions on $\mathbf{x}$ and is optimized to allocate reasoning 
effort based on input difficulty;

\textbf{(2)~Self-evaluation}—$\mathcal{P}(\mathbf{r}, \mathbf{x})$ 
captures internal feedback signals (e.g., confidence, intermediate 
rewards) that guide reasoning decisions;

\textbf{(3)~Cost-aware optimization}—the term 
$\lambda \cdot \mathcal{C}(\mathbf{r})$ explicitly balances 
correctness and efficiency.

These properties are realized through training-based methods 
(\S\ref{sec:training-based}) that optimize $\pi_\theta$ during 
training, or training-free methods (\S\ref{sec:training-free}) that apply dynamic control at inference. This distinction forms 
the foundation for our taxonomy in \S\ref{sec:taxonomy}.

\section{Taxonomy of Adaptive Reasoning}
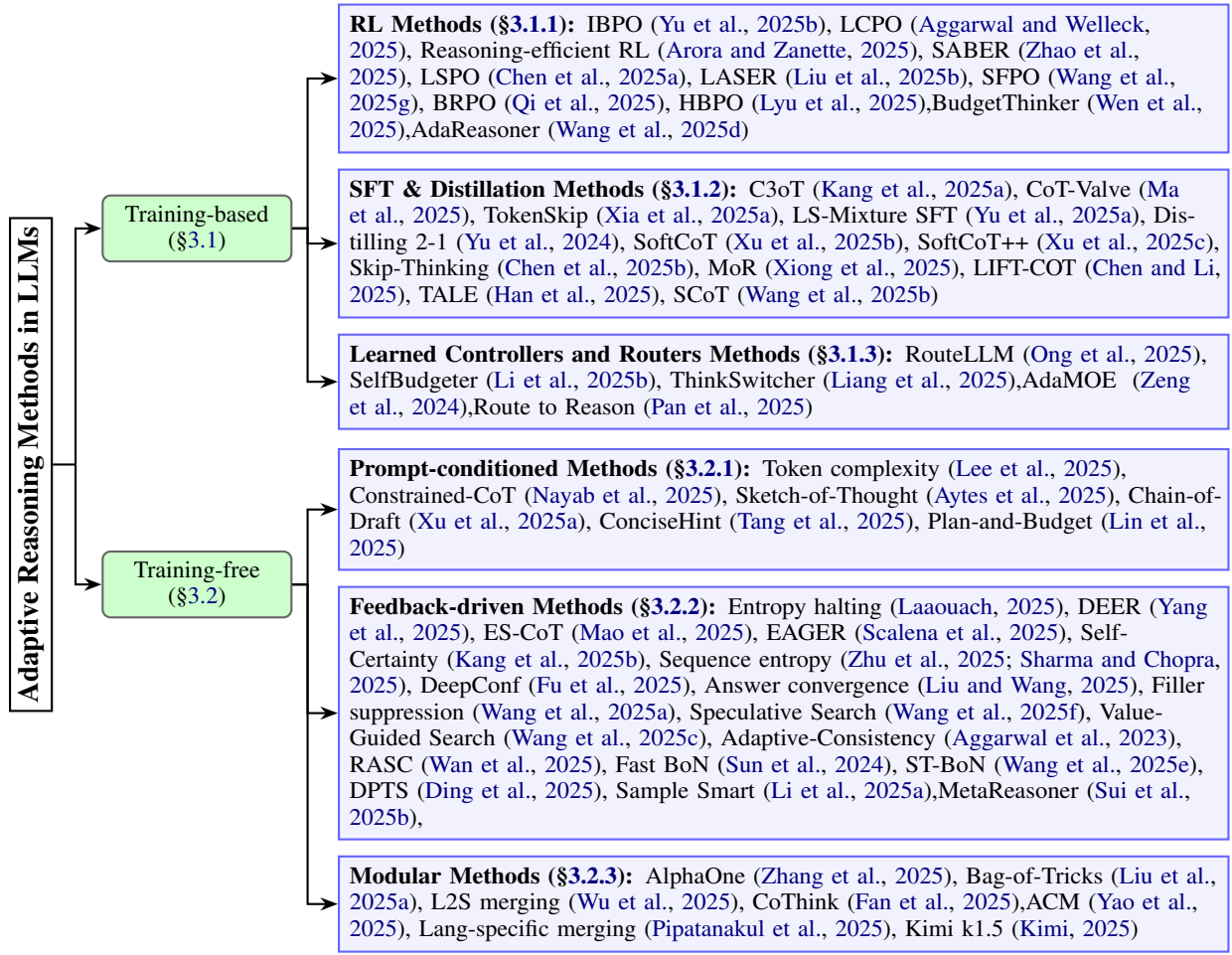
\begin{figure*}[t]
\centering
\begin{tikzpicture}[
    node distance = 0.4cm and 0.5cm,
    method/.style={
        rectangle, 
        draw=black!60, 
        fill=green!20, 
        thick,
        minimum width=2.5cm,
        minimum height=0.8cm,
        align=center,
        rounded corners=3pt,
        font=\small
    },
    content/.style={
        rectangle,
        draw=blue!60,
        fill=blue!5,
        thick,
        text width=11.5cm,
        minimum height=0.6cm,
        align=left,
        inner sep=4pt,
        font=\small
    },
    rootbox/.style={
        rectangle,
        draw=black,
        fill=white,
        thick,
        inner sep=3pt,
        font=\normalsize\bfseries
    },
    arrow/.style={
        -Stealth,
        thick
    }
]
\node[rootbox] (rootlabel) at (-2, 0) {\rotatebox{90}{Adaptive Reasoning Methods in LLMs}};
\node[method] (train) at (0.2, 3.15) {Training-based\\(\S\ref{sec:training-based})};
\node[method] (free) at (0.2, -1.6) {Training-free\\(\S\ref{sec:training-free})};
\node[content, right=0.6cm of train, yshift=2cm] (rl) {
    \textbf{RL Methods (\S\ref{sec:rl}):} IBPO~\cite{Yu2025IBPO}, LCPO~\cite{Aggarwal2025LCPO}, Reasoning-efficient RL~\cite{Arora2025EfficientRL}, SABER~\cite{Zhao2025SABER}, LSPO~\cite{Chen2025LSPO}, LASER~\cite{liu2025learnreasonefficientlyadaptive}, SFPO~\cite{Li2025SFPO}, BRPO~\cite{qi2025optimizinganytimereasoningbudget}, HBPO~\cite{lyu2025hierarchicalbudgetpolicyoptimization},BudgetThinker~\cite{wen2025budgetthinker},AdaReasoner~\cite{Tan2025AdaReasoner}
};
\node[content, below=0.2cm of rl] (sft) {
    \textbf{SFT \& Distillation Methods (\S\ref{sft}):} C3oT~\cite{kang2025c3ot}, CoT-Valve~\cite{Ma2025CoTValve}, TokenSkip~\cite{Xia2025TokenSkip}, LS-Mixture SFT~\cite{Yu2025LSMixtureSFT}, Distilling 2-1~\cite{Zhang2024DistillingTwoOne}, SoftCoT~\cite{Xu2025SoftCoT}, SoftCoT++~\cite{Xu2025SoftCoTpp}, Skip-Thinking~\cite{chen2025skipthinkingchunkwisechainofthoughtdistillation}, MoR~\cite{xiong2025mixturereasoningsteachlarge}, LIFT-COT~\cite{electronics14081662}, TALE~\cite{Han2025TALE},  SCoT~\cite{wang2025efficientreasoningllmsspeculative}
};
\node[content, below=0.2cm of sft] (router) {
    \textbf{Learned Controllers and Routers Methods (\S\ref{sec:controller routers}):} RouteLLM~\cite{ong2024routellm}, SelfBudgeter~\cite{Li2025SelfBudgeter}, ThinkSwitcher~\cite{Liang2025ThinkSwitcher},AdaMOE ~\cite{zeng2024adamoe},Route to Reason~\citep{pan2025routereasonadaptiverouting}
};

\node[content, right=0.6cm of free, yshift=1cm] (prompt) {
    \textbf{Prompt-conditioned Methods (\S\ref{sec:prompt-conditioning}):} Token complexity~\cite{Lee2025TokenComplexity}, Constrained-CoT~\cite{Nayab2025ConciseThoughts}, Sketch-of-Thought~\cite{aytes2025sketchofthoughtefficientllmreasoning}, Chain-of-Draft~\cite{Xu2025ChainOfDraft}, ConciseHint~\cite{Tang2025ConciseHint}, Plan-and-Budget~\cite{Lin2025PlanBudget}
};
\node[content, below=0.2cm of prompt] (feedback) {
    \textbf{Feedback-driven Methods (\S\ref{sec:feedback}):} Entropy halting~\cite{laaouach2025halt}, DEER~\cite{Yang2025DEER}, ES-CoT~\cite{mao2025earlystoppingchainofthoughtslarge}, EAGER~\cite{Scalena2025EAGER}, Self-Certainty~\cite{Kang2025SelfCertainty}, Sequence entropy~\cite{zhu2025uncertainty,sharma2025think}, DeepConf~\cite{Fu2025DeepConf}, Answer convergence~\cite{liu2025answerconvergence}, Filler suppression~\cite{wang2025nowait}, Speculative Search~\cite{wang2025specreasoning}, Value-Guided Search~\cite{wang2025valueguided}, Adaptive-Consistency~\cite{aggarwal2023letssamplestepstep}, RASC~\cite{wan2025reasoningawareselfconsistencyleveraging}, Fast BoN~\cite{Sun2024FastBoN}, ST-BoN~\cite{Wang2025STBoN}, DPTS~\cite{Ding2025DPTS}, Sample Smart~\cite{Li2025SampleSmartNotHard},MetaReasoner~\cite{sui2025metareasonerdynamicguidanceoptimized}, 
};
\node[content, below=0.2cm of feedback] (modular) {
    \textbf{Modular Methods (\S\ref{sec:modular}):}  AlphaOne~\cite{Zhang2025AlphaOne}, Bag-of-Tricks~\cite{liu2025bagtricksinferencetimecomputation}, L2S merging~\cite{Wu2025L2S}, CoThink~\cite{fan2025cothink},ACM~\cite{Yao2025ACM}, Lang-specific merging~\cite{Pipatanakul2025LangSpecificMerge}, Kimi k1.5~\cite{TeamKimi2025Kimi15}
};
\draw[thick] (rootlabel.east) -- ++(0.3,0) coordinate (junction);
\draw[arrow] (junction) |- (train.west);
\draw[arrow] (junction) |- (free.west);
\draw[arrow] (train.east) -- ++(0.2,0) |- (rl.west);
\draw[arrow] (train.east) -- ++(0.2,0) |- (sft.west);
\draw[arrow] (train.east) -- ++(0.2,0) |- (router.west);
\draw[arrow] (free.east) -- ++(0.2,0) |- (prompt.west);
\draw[arrow] (free.east) -- ++(0.2,0) |- (feedback.west);
\draw[arrow] (free.east) -- ++(0.2,0) |- (modular.west);
\end{tikzpicture}
\caption{Taxonomy of adaptive reasoning methods in LLMs.}
\label{fig:taxonomy}
\end{figure*}
\label{sec:taxonomy}

\subsection{Training-based Adaptive Reasoning}
\label{sec:training-based}

Training-based adaptive reasoning methods aim to endow models with the ability to regulate their reasoning depth and computational cost \emph{during training}.
Instead of applying external halting or sampling rules, these methods modify learning objectives or data organization so that the model internalizes
a sense of when to reason longer, when to compress, and how to allocate computational resources based on instance difficulty or uncertainty.
This section reviews five main families of approaches that operationalize such adaptivity at training time.

\subsubsection{Reinforcement Learning Methods}
\label{sec:rl}
Reinforcement-learning (RL)–based approaches integrate reasoning accuracy and computational efficiency into a unified optimization objective.
IBPO \citep{Yu2025IBPO} formulates an inference-budget–constrained policy optimization problem where the policy learns to allocate
larger token budgets to harder questions and smaller ones to easier instances.
LCPO \citep{Aggarwal2025LCPO} directly adds a length-adherence(i.e., a reward that penalizes deviations from the target reasoning length) reward term to policy gradients,
enabling the agent to match target reasoning lengths and revealing the emergence of
short-reasoning models that preserve logical fidelity.
\citep{Arora2025EfficientRL} propose a reasoning-efficient RL framework
that shapes rewards to favor concise yet correct reasoning trajectories.
SABER \citep{Zhao2025SABER} introduces switchable and balanced reward schedules to train models under multiple budget tiers.
LSPO \citep{Chen2025LSPO} develops a length-aware data sampling strategy that dynamically filters training batches based on response length (e.g., retaining the shortest and longest) to stabilize training.
Other variants include LASER \citep{liu2025learnreasonefficientlyadaptive}, which introduces adaptive, difficulty-aware reward shaping, penalizing lengthy reasoning more heavily on simple tasks, and Slow-Fast Policy Optimization (SFPO) \citep{Li2025SFPO} structures each update into fast, reposition, and slow phases to reduce gradient noise and stabilize convergence. Multi-objective approaches such as 
{BRPO} \citep{qi2025optimizinganytimereasoningbudget} optimizes anytime reasoning via budget-sampled dense rewards and a variance-reduced policy gradient, and
{HBPO} \citep{lyu2025hierarchicalbudgetpolicyoptimization} structures exploration across hierarchical token-budget groups to preserve capability while reducing average tokens.
Together, these works demonstrate how reinforcement learning can encode budget awareness and trade-offs directly into learned reasoning policies through principled reward shaping and constraint design.
Other variants include
BudgetThinker \citep{wen2025budgetthinker} introduces dynamic control tokens whose embeddings encode remaining compute, enabling the model to adjust reasoning depth continuously during decoding, while uniquely combining an RL-based reward function with explicit control tokens, bridging the gap between purely implicit policy optimization and supervised control. AdaReasoner \citep{Tan2025AdaReasoner} learns an adaptive policy via reinforcement learning to dynamically configure reasoning parameters (e.g., temperature, step length, prompt type) for different tasks

\subsubsection{Supervised Fine-tuning and Distillation Methods}
\label{sft}
A complementary line of work leverages supervised fine-tuning (SFT) or distillation to
teach models to produce shorter yet equally valid reasoning chains.
C3oT \citep{kang2025c3ot} jointly trains on long–short pairs of chain-of-thought (CoT) exemplars,
allowing conditional generation of compact rationales without accuracy loss.
CoT-Valve \citep{Ma2025CoTValve} identifies a controllable ``length direction'' in parameter space and tunes it with paired long/short data for elastic reasoning compression.
TokenSkip \citep{Xia2025TokenSkip} prunes semantically redundant tokens and retrains the model under different compression ratios to learn explicit compression control.
LS-Mixture SFT \citep{Yu2025LSMixtureSFT} integrates long and short CoT data during fine-tuning,
enabling adaptive reasoning depth across instances.
Distilling 2-1 \citep{Zhang2024DistillingTwoOne} removes explicit reasoning chains during distillation while retaining correctness, encouraging implicit short reasoning behaviors.
SoftCoT \citep{Xu2025SoftCoT} and SoftCoT++ \citep{Xu2025SoftCoTpp} represent thought processes as continuous latent vectors, allowing smooth trajectory distillation that generalizes across prompts.
Skip-Thinking \citep{chen2025skipthinkingchunkwisechainofthoughtdistillation} conducts chunk-wise distillation to mitigate gradient saturation in long CoT training. {Mixture of Reasonings (MoR)} \citep{xiong2025mixturereasoningsteachlarge} jointly trains LLMs on multiple reasoning strategies (e.g., logical, commonsense, numerical) through multi-style CoT supervision, enabling implicit adaptive reasoning style selection without explicit routing.
TALE \citep{Han2025TALE} performs token-budget–aware prompt tuning that learns to predict minimal feasible token counts per instance.
Length-Instruction Fine-Tuning with Chain-of-Thought (LIFT-COT) \citep{electronics14081662}
extends standard supervision with explicit length labels,
enabling the model to adhere to user-specified reasoning length or latency constraints
while maintaining coherence in its generated rationales. SCoT \citep{wang2025efficientreasoningllmsspeculative} introduces a speculative chain-of-thought framework where a small draft model rapidly generates multiple reasoning trajectories, and a large target model selects or corrects them, achieving up to 3× faster inference with near-target-level accuracy. 
Together, these supervised and distillation-based methods internalize compression control within model parameters, achieving adaptive reasoning efficiency without reinforcement objectives.
\subsubsection{Learned Controllers and Routers}
\label{sec:controller routers}

Another line of work introduces auxiliary controllers or routing modules trained jointly or sequentially with the base model to decide
which reasoning path or sub-model to activate.
RouteLLM \citep{ong2024routellm} learns a lightweight router from preference data to dispatch easy inputs to cheaper small models and
difficult ones to larger models, reducing total compute without degrading accuracy.
SelfBudgeter \citep{Li2025SelfBudgeter} trains a two-stage system that first predicts the required reasoning budget and then enforces it through
gradient-penalized reinforcement optimization. {ThinkSwitcher}~\citep{Liang2025ThinkSwitcher} extends this idea by training a supervised switching module that decides between concise and elaborate reasoning trajectories based on predicted task difficulty.  In Mixture-of-Experts (MoE) models, AdaMOE~\citep{zeng2024adamoe} achieves token-adaptive computation by introducing "null experts" that consume zero FLOPs. A load-balancing loss is used during training to teach the model's router to adaptively send tokens to either true experts for computation or null experts to skip computation, thereby dynamically allocating resources at the token level.  Route-To-Reason (RTR)~\citep{pan2025routereasonadaptiverouting} learns to predict accuracy–cost trade-offs across model–strategy pairs and dynamically routes each query to the optimal expert combination under a compute budget. Collectively, these works embed adaptivity at the system-decision level and bridge symbolic routing with learned compute control.

\subsection{Training-free Adaptive Reasoning}
\label{sec:training-free}
This section overviews training-free adaptive reasoning, where model parameters remain frozen and adaptivity arises from inference-time control. It introduces three categories: prompt-conditioned, feedback-driven, and modular methods.

Training-free adaptive reasoning denotes inference-time mechanisms that modulate the \emph{depth, breadth, or structure} of reasoning without any parameter update.
Formally, a frozen model $f_\theta$ generates a reasoning trajectory $r=(t_1,\dots,t_n)$ governed by a control policy $\pi(x,t_i)$ that dynamically decides continuation or termination based on input difficulty or internal uncertainty.
Unlike static efficiency control, these approaches allocate computation adaptively for each instance, emphasizing resource-aware and self-regulated inference
\citep{Sui2025StopOverthinking,alomrani2025reasoningbudgetsurveyadaptive,zhu2025conciseadaptivethinkinglarge}.

\subsubsection{Prompt-conditioned Methods}
\label{sec:prompt-conditioning}

\textbf{Explicit token- and instruction-level control.}
Prompt-conditioned methods introduce explicit instructions, control tokens, or concise templates that determine reasoning length at test time.
They embody \emph{external} adaptivity—computation is modulated through textual conditions rather than learned signals \citep{cai2025does}.
\citet{Lee2025TokenComplexity} quantifies token complexity to estimate minimal reasoning spans per instance.
Contraied-CoT \citep{Nayab2025ConciseThoughts} employs templates that prioritize conciseness, and enforce concise reasoning by setting explicit length constraints (e.g., '100 words'). Sketch-of-Thought \citep{aytes2025sketchofthoughtefficientllmreasoning}
 introduces cognitive-inspired sketching to generate compact, structured reasoning trajectories.
 {Chain-of-Draft} (CoD) \citep{Xu2025ChainOfDraft} introduces a “draft-first” reasoning style: the model writes short, essential reasoning steps (at most ~5 words) per turn, thus minimizing verbosity while preserving correctness. {ConciseHint} \citep{Tang2025ConciseHint} integrates continuous hints or contextual anchors for difficulty-aware reasoning.
Instruction-level prompt schemes such as Plan-and-Budget \citep{Lin2025PlanBudget}
introduce explicit planning and budgeting instructions during inference,
enabling the model to allocate reasoning depth and token usage adaptively
across sub-questions.

Together, these approaches treat reasoning efficiency as a prompt-engineering problem with {explicit, symbolic, yet parameter-free} adaptivity.

\subsubsection{Feedback-driven Methods}
\label{sec:feedback}

\textbf{Uncertainty- and entropy-based halting.}
Feedback-driven adaptation relies on internal feedback such as entropy, confidence, or semantic closure to determine when reasoning should stop or expand.
\citet{laaouach2025halt} defines entropy-threshold halting, where decoding terminates once uncertainty drops below a pre-defined threshold.
\citet{Yang2025DEER} introduces a feedback-driven early exit mechanism that monitors token-level confidence during reasoning to adaptively halt generation once sufficient certainty is reached.
\citet{mao2025earlystoppingchainofthoughtslarge} 
employs a run-jump test to detect answer convergence, terminating the reasoning process when a sufficient number of consecutive steps yield identical intermediate answers.
{EAGER} \citep{Scalena2025EAGER} leverages token-level entropy to branch into multiple reasoning paths only when high uncertainty is detected. {Self-Certainty} \citep{Kang2025SelfCertainty} proposes a metric that evaluates answer quality from the model’s own output probabilities, offering a reward-model-free substitute for Best-of-N selection.
Sequence-level entropy 
\citep{zhu2025uncertainty,sharma2025think} and confidence-based filtering methods such as DeepConf\citep{Fu2025DeepConf} further refine this paradigm. 

\textbf{Verifier- and constraint-guided feedback.}
Beyond internal entropy and confidence, some approaches externalize feedback via verifiers or structural constraints.
\citet{liu2025answerconvergence} detect \emph{answer convergence} across partial chains and stop early when consistency stabilizes.
\citet{wang2025nowait} prune filler \emph{thinking} tokens by suppressing keywords (e.g., ``Wait'', ``Hmm'') during decoding.
Speculative or constraint-based schemes such as {Speculative Search} \citep{wang2025specreasoning} and {Value-Guided Search} \citep{wang2025valueguided} use acceptance thresholds or value models to admit only high-quality thoughts under limited budgets.

MetaReasoner~\citep{sui2025metareasonerdynamicguidanceoptimized} employs a contextual bandit controller that dynamically adjusts reasoning depth during inference based on external validation signals, exemplifying feedback-driven adaptation without reinforcement fine-tuning.
These methods remain feedback-driven in essence: reasoning continues or halts based on real-time validation rather than fixed templates.

\textbf{Consistency- and sampling-based adaptation.}
A parallel line of work focuses on reasoning breadth—how many rationales to sample before aggregation.
{Adaptive-Consistency} \citep{aggarwal2023letssamplestepstep} and {RASC} \citep{wan2025reasoningawareselfconsistencyleveraging} estimate consensus and halt when majority agreement stabilizes.

Best-of-$N$ optimization is improved by {Fast BoN} \citep{Sun2024FastBoN}, which employs efficient sampling algorithms to generate high-scoring responses,and {ST-BoN} \citep{Wang2025STBoN}, which leverages early sampling consistency to identify and truncate suboptimal reasoning paths before full generation, while {DPTS} \citep{Ding2025DPTS} adaptively prunes search branches.
Correctness-first sampling \citep{Li2025SampleSmartNotHard}, also known as Sample Smart, further balance accuracy and latency, refining adaptive breadth control for reasoning LLMs.

Collectively, these methods realize an {internal self-regulation}that reasoning is guided by feedback loops rather than pre-specified templates.

\subsubsection{Modular Methods}
\label{sec:modular}

This category covers systems that combine or reuse pre-trained reasoning models through composition, merging, or ensembling.
To clarify the boundary with \ref{sec:controller routers}, we \emph{treat any static, no-gradient weight interpolation or parameter merging} (e.g., long–short or activation-guided merging) 
as \textbf{training-free modular adaptation}.
In contrast, frameworks that involve training new gating, routing, or fusion modules are considered \textbf{learned controllers or routers} (\ref{sec:controller routers}).

\textbf{Pipeline-based adaptive inference.}
At the system level, modular methods dynamically allocate reasoning compute through a plan–budget–execute pipeline.

Although implemented through decoding-level control rather than explicit architectural modules,
{AlphaOne} \citep{Zhang2025AlphaOne}
employs pacing tokens to regulate cognitive tempo across short-, long-, and tool-augmented reasoning phases.
This design mimics modular behavior by dynamically allocating computation among distinct reasoning modes,
bridging concise and deliberative thinking within a unified system. CoThink \citep{fan2025cothink} introduces a training-free modular pipeline where an instruct model first drafts a concise reasoning outline, and a thinking model expands it, reducing token usage while maintaining accuracy.
Industrial frameworks such as {Bag-of-Tricks for Inference-Time Compute} \citep{liu2025bagtricksinferencetimecomputation} 
systematically benchmarks training-free modular techniques that combine generation, verification, and confidence-weighted selection to improve inference-time reasoning efficiency. These systems demonstrate how architectural modularity induces reasoning flexibility without retraining. 

\textbf{Model-merging and ensemble-based adaptation.}

A complementary branch combines pre-trained reasoning modules by parameter interpolation or selective ensembling.
Long-to-short reasoning via model merging(L2S merging) \citep{Wu2025L2S} reduces average reasoning length while preserving accuracy, and
activation-informed merging(ACM)\citep{Yao2025ACM} learns layer-wise coefficients from activation mutual information.
For language-specific settings, model merging can transfer reasoning ability from a strong “reasoner” into a target-language model without retraining \citep{Pipatanakul2025LangSpecificMerge}.
Commercial deployments such as Kimi k1.5 also report hybrid short/long reasoning variants obtained by merging long-CoT and short-CoT specialists \citep{TeamKimi2025Kimi15}.
These practices share the same anchor that {adaptive compute distribution across fixed modules} via weight interpolation or expert ensembling.
\section{Conclusion and Open Challenges}
This survey reframes reasoning in LLMs through the lens of adaptivity, defining it as the dynamic modulation of reasoning resources based on input-dependent complexity. We formalize this via classical cognitive paradigms and a unified taxonomy of training-based and training-free mechanisms. 

However, as the field shifts toward test-time scaling exemplified by models like OpenAI o1 \cite{jaech2024openai} and DeepSeek-R1 \cite{guo2025deepseek}, new challenges emerge. A primary concern is the reliability of long-chain reasoning; while scaling inference computing can unlock complex problem-solving, it frequently leads to ``reasoning drift,'' where minor errors accumulate into hallucinations. Future adaptivity mechanisms must therefore transition from simple halting to active error-aware backtracking and dynamic pruning of reasoning paths. Furthermore, the meta-reasoning overhead remains a bottleneck; adaptation signals must be lightweight, potentially utilizing embedding-level difficulty estimations, to ensure that the cost of deciding {how} to reason does not offset the efficiency gains. Crucially, extending these mechanisms to dynamic and embodied systems represents a major frontier. Recent work such as Ares \cite{yang2026ares} and RARRL \cite{liu2026should} demonstrate that agents must autonomously balance high-level cognition with execution constraints, while others adapt via continuous environmental feedback \cite{kurokitime}. Finally, as adaptive reasoning extends into creative and exploratory domains, the community requires more nuanced metrics that go beyond exact-match accuracy to capture the marginal utility and semantic richness of test-time compute.

\section{Limitations}
This survey inevitably faces two limitations.
First, due to space constraints, we focus on representative adaptive reasoning methods and do not exhaustively cover all recent variants, especially those emerging in multimodal or agentic settings. Our taxonomy emphasizes conceptual clarity over completeness, which may overlook some niche but relevant techniques.
Second, the field of adaptive reasoning is evolving rapidly and new paradigms such as self-improving reasoning, adaptive reflection, and meta-evaluation appear almost monthly. Consequently, future developments may shift or refine the taxonomy presented here.
\bibliography{custom}

\end{document}